\documentclass[lettersize,journal]{IEEEtran}
\usepackage{amsmath,amsfonts}
\usepackage{algorithmic}
\usepackage{algorithm}
\usepackage{array}
\usepackage[caption=false,font=normalsize,labelfont=sf,textfont=sf]{subfig}
\usepackage{textcomp}
\usepackage{stfloats}
\usepackage{url}
\usepackage{verbatim}
\usepackage{graphicx}
\usepackage{cite}

\usepackage{tabu}                      
\usepackage{booktabs}                  
\usepackage{lipsum}                    
\usepackage{mwe}

\usepackage{mathptmx}
\usepackage{amsmath}
\usepackage{multirow}
\usepackage{graphicx}
\usepackage{enumitem}
\usepackage{wrapfig}
\usepackage{xspace}

\usepackage{amssymb}
\usepackage{pifont}
\usepackage{hyperref}

\newcommand{\xmark}{\ding{55}}%
\newcommand{\ie}{i.e.,\xspace}
\newcommand{\eg}{e.g.,\xspace}

\newcommand{\ra}[1]{\renewcommand{\arraystretch}{#1}}
\PassOptionsToPackage{svgnames}{xcolor}
\usepackage{xcolor}

\definecolor{NavyBlue}{RGB}{0,0,0}
\definecolor{revBlue}{RGB}{0,0,0}
\definecolor{DodgerBlue}{RGB}{0, 0, 0}
\definecolor{red}{RGB}{0, 0, 0}
\definecolor{hlBlue}{RGB}{0,0,0}

\newcommand{\hl}[1]{\textcolor{hlBlue}{#1}}

\begin{document}

\title{Generalization of CNNs on \\ Relational Reasoning with Bar Charts}

\author{Zhenxing Cui\textsuperscript{*},
	Lu Chen\textsuperscript{*},
    Yunhai Wang,
	Daniel Haehn,
    Yong Wang,
	and Hanspeter Pfister~\IEEEmembership{IEEE Fellow}

\IEEEcompsocitemizethanks{
    \IEEEcompsocthanksitem * Zhenxing Cui \& Lu Chen contribute equally to this work, and Yunhai Wang is the corresponding author.
	\IEEEcompsocthanksitem Zhenxing Cui is with the School of Computer Science and Technology, Shandong University, China.	E-mail: zhenxingcui@mail.sdu.edu.cn
	\IEEEcompsocthanksitem Lu Chen is with the State Key Lab of CAD\&CG, Zhejiang University, China. E-mail: lu.chen@zju.edu.cn
    \IEEEcompsocthanksitem Yunhai Wang is with the School of Information, Renmin University of China, China. E-mail: wang.yh@ruc.edu.cn
    \IEEEcompsocthanksitem Daniel Haehn is with the College of Science and Mathematics, University of Massachusetts Boston, USA. E-mail: haehn@cs.umb.edu
	\IEEEcompsocthanksitem Yong Wang is with the College of Computing and Data Science, Nanyang Technological University, Singapore. E-mail: yong-wang@ntu.edu.sg
	\IEEEcompsocthanksitem Hanspeter Pfister is with the School of Engineering and Applied Sciences, Harvard University, USA. E-mail: pfister@seas.harvard.edu
    
}

\thanks{Manuscript received May 22, 2023; revised September 15, 2024.}}

\maketitle

\begin{abstract}

This paper presents a systematic study of the generalization of convolutional neural networks (CNNs) and humans on relational reasoning tasks with bar charts. We first revisit previous experiments on graphical perception and update the benchmark performance of CNNs. We then test the generalization performance of CNNs on a classic relational reasoning task: estimating bar length ratios in a bar chart, by progressively perturbing the standard visualizations. We further conduct a user study to compare the performance of CNNs and humans. Our results show that CNNs outperform humans only when the training and test data have the same visual encodings. Otherwise, they may perform worse. We also find that CNNs are sensitive to perturbations in various visual encodings, regardless of their relevance to the target bars. Yet, humans are mainly influenced by bar lengths. Our study suggests that robust relational reasoning with visualizations is challenging for CNNs. Improving CNNs’ generalization performance may require training them to better recognize task-related visual properties.

\end{abstract}


\section{Introduction}

\IEEEPARstart{D}{EEP} neural networks, especially convolutional neural networks (CNNs), are increasingly being adopted in the visualization community for many tasks such as visual question answering~\cite{kahou2017figureqa,kafle2018dvqa},  automatic visualization design~\cite{chen2019towards}, and chart captioning~\cite{kanthara2022chart,mahinpei2022linecap}.
Despite their widespread use, the crucial question of how well these models generalize to previously unseen visualizations remains less explored.
Understanding and enhancing this generalization ability is crucial for the real-world deployment of CNNs.

Graphical perception~\cite{cleveland1984graphical} refers to the human ability to decode visually encoded quantities in visualizations.
It plays a foundational role in understanding the relations between visual elements, such as the bar length ratios in bar charts.
Seminal work by Cleveland and McGill~\cite{cleveland1984graphical}, along with subsequent studies~\cite{heer2010crowdsourcing,talbot2014four} highlighted human proficiency with aligned bars over stacked configurations.
Recent findings by Haehn et al.~\cite{haehn2019evaluating} indicate that CNNs do not match human accuracy in interpreting bar charts, raising questions about the discrepancy in graphical perception between CNNs and humans.

Our research aims to address two primary shortcomings of the current evaluation of CNNs in graphical perception.
First, a common limitation of existing studies is their reliance on oversimplified stimuli that fail to capture the complexity of standard visualizations. For example, the stimuli used by Haehn et al.~\cite{haehn2019evaluating} were all black-and-white images devoid of axes, legends, or titles, which are essential chart components according to the Grammar of Graphics~\cite{wilkinson2012grammar}. 
We address this limitation by evaluating the performance of CNNs and humans using more realistic bar charts that better reflect standard visualization design.

Furthermore, although CNNs have demonstrated remarkable performance in object recognition, their robustness to visual changes is limited, as highlighted by Geirhos et al.~\cite{geirhos2018generalisation}.
We hypothesize this limited robustness extends to relational reasoning tasks, particularly in the context of commonly used bar charts, which have not been thoroughly investigated.
Our study seeks to systematically evaluate the generalization capabilities of CNNs in visual relational reasoning, focusing on their adaptability to novel charts with variations in key visualization parameters such as stroke width, title position, and background color, which reflect a wide range of design choices driven by individual preferences.

In this paper, we present a comprehensive study of the graphical perception capabilities of CNNs and humans. We focus on the ratio estimation task in bar charts and aim to answer the following research questions:
\begin{enumerate}[label=(\roman*)]
\vspace{-1mm}
\item How well do CNNs perform and generalize on standard visualizations that include all necessary components specified by a visualization grammar?
\item How robust are CNNs to perturbations of different visual properties in standard visualizations?
\item What are the major differences between CNNs and humans in visual relational reasoning?
\vspace{-1mm}
\end{enumerate}
Our answers to these questions will help us better understand the potential generalization ability of CNNs and humans and provide guidelines for training and designing robust CNNs for relational reasoning tasks involving data visualizations.

We first revisit the experiments conducted by Haehn et al.~\cite{haehn2019evaluating} and find that a CNN model trained with the identified (near) optimal configurations can estimate bar length ratios more accurately than humans when the training and test data have similar visual encodings. Moreover, this CNN model can also evaluate bar charts with an expanded parametric space, such as an increased number of possible ratios in ratio estimation tasks.

Next, we conduct a structured generalization analysis of CNNs and humans on standard visualizations with a wider range of visual appearances, synthesized using Vega-Lite~\cite{satyanarayan2016vega}, a high-level visualization grammar. We evaluate the performance of CNNs on visualizations without and with progressive perturbations of various visual parameters, such as title position, background color, bar width, and bar length.

We also conduct a user study to measure human performance under the same conditions. The results show that: 
(i) CNNs perform better than humans in relational reasoning on bar charts when the testing data mirrors the visual encodings of the training data. However, their performance can quickly decline below human levels on charts that incorporate perturbations in certain visual encodings; and (ii) CNNs exhibit less robustness to perturbations of various visual encodings, even those unrelated to the target visual marks. In contrast, humans are mainly influenced by the lengths of the bars in the chart.

Our study suggests that humans and CNNs may use different reasoning processes, with humans primarily focusing on specific parameters of the target bars while ignoring other irrelevant chart details. To understand why CNNs fail to generalize to perturbed visualizations, we visualize the important regions used by CNNs for estimating the ratios using gradient-weighted class activation mapping (Grad-CAM)~\cite{selvaraju2017grad} and find that the regions of target bars are rarely used as the key evidence by CNNs trained on this task. However, by providing enough guidance of the target regions, their ability to generalize to perturbed visualizations can be improved. Even so, CNNs still do not use the lengths of the bars as their sole basis for estimating ratios.

In summary, this work makes several novel contributions. 
We systematically study the generalization performance of humans and CNNs on a large number of standard visualizations synthesized according to the grammar of graphics. We demonstrate that appropriately tuned hyperparameters can enable most CNN models to meet or exceed human performance on relational reasoning tasks when the training and test data have similar visual encodings. Finally, we show that providing ground-truth target region information can improve the generalization performance of CNNs on graphical perception tasks.
Data, source code, study materials, and additional analyses are included in our supplemental material and can be found at \url{https://github.com/Ideas-Laboratory/Graphical-Perception}.

\section{Related Work}

\subsection{Graphical Perception}

The concept of graphical perception was introduced by Cleveland and McGill to describe the visual decoding of information encoded in graphs~\cite{cleveland1984graphical,cleveland1985graphical}. By conducting a series of elementary perceptual experiments on different visual encodings, Cleveland and McGill evaluated the perceptual difficulty of different visual encodings for humans. Their work was later replicated by a number of follow-up studies, such as Heer and Bostock's evaluation of human graphical perception on Mechanical Turk~\cite{heer2010crowdsourcing} and Harrison et al.'s reproduction of the Cleveland-McGill experiments with a focus on the influence of affective priming~\cite{harrison2013influencing}.
These studies have established a foundation for understanding human capabilities in interpreting information from visualizations. 
Building on this body of knowledge, Zeng et al.~\cite{zeng2024too} leverage insights from graphical perception to guide visualization recommendations.

Recently, researchers have focused on the building blocks of visualizations, such as position, length, and angle estimation, and have explored how well neural networks can perform these elementary perceptual tasks. Haehn et al.\cite{haehn2019evaluating} investigated whether off-the-shelf CNNs can predict human responses to graphical perception building blocks by replicating Cleveland and McGill's experiments~\cite{cleveland1984graphical} (see Section~\ref{sec:background}). Haleem et al.~\cite{haleem2019evaluating} explored the potential of using a handcrafted CNN to estimate the readability of graph visualizations and found that machine graphical perception has promise for graph visualizations. Giovannangeli et al.~\cite{giovannangeli2020toward} found that current CNNs are able to replicate previously published studies of human graph perception while still exhibiting generalization limitations.

However, all of these studies evaluate the generalization ability of CNNs using data generated from highly controlled settings, which are far from real-world scenarios. We argue that the evaluation should be done with standard visualizations created according to the grammar of graphics~\cite{wilkinson2012grammar}. Otherwise, the results may not be representative of meaningful generalization, and such controlled settings may limit the usefulness of CNNs. In addition, we aim to investigate the robustness of CNNs against varying visual properties and compare them with human performance.

\subsection{CNNs for Visualization Generation and Analysis}

Inspired by the great success of CNNs in general computer vision and graphics tasks, there has been an increasing interest in applying them
to data visualizations. The studies can be categorized into two groups: visualization generation and visualization analysis.

In visualization generation, CNNs are employed to ease the creation of visualizations and encode additional information into  visualizations~\cite{zhang2020viscode,fu2020chartem}.
VizML~\cite{hu2019vizml} utilizes a three-layer CNN to automatically recommend a suitable visual encoding of standard charts, including axis specifications and chart types.
Chen et al.~\cite{chen2019towards} leverage the ResNeXt architecture~\cite{xie2017aggregated} to decompose timeline infographics and further construct infographics with a similar design style for new input data.

Prior studies on visualization analysis focus on training CNNs to understand data visualizations. A representative task is chart Question Answering (QA)~\cite{kahou2017figureqa,methani2019plotqa}.
Since simple CNNs have difficulties solving the visual relation reasoning problem, Kafle et al.~\cite{kafle2018dvqa} extend Stacked Attention Networks~\cite{yang2016stacked} with dynamic encoding to enhance the network's understanding of bar charts.
VisQA~\cite{kim2020answering} adopts a template-based approach to generate explanations along with automatic question answering. 
CNNs have also been applied for various other visualization analysis tasks,
such as visualization retrieval~\cite{saleh2015learning}, outlier identification~\cite{Giovannangeli2021perception,Giovannangeli2022Color}, design pattern analysis~\cite{yuan2021deepcolormap}, and chart ensemble exploration~\cite{dai2018chart,zhao2020chartseer}. 

All the above techniques intrinsically rely on the graphical perception capability of CNNs.
In this work, we systematically evaluate how different CNN architectures perform in graphical perception tasks while providing implications for training and designing more powerful CNN models for visualization tasks.

\subsection{CNN Generalization for Visualizations}

CNNs have shown remarkable generalization abilities on data in which training  and test data have similar distributions, outperforming humans in various tasks like image classification and object recognition.
Despite these successes, their generalization performance degrades significantly when faced with out-of-distribution (OOD) test data~\cite{geirhos2018generalisation,geirhos2018imagenet,hendrycks2018benchmarking,hendrycks2021many}.

Only a few studies have been conducted regarding the OOD performance of CNNs for visualization techniques with small perturbations in the input.
Lopes and Brodlie~\cite{lopes2003improving} evaluated the robustness of isosurfaces by observing how the continuity varies with respect to changes in the data or isosurface level.
Correll et al.~\cite{correll2019looks} show that a small perturbation of visual parameters might obscure the important data patterns. 
Wang et al.~\cite{wang2019improving} examined the robustness of scagnostic measures by conducting a set of empirical studies. 
Yang et al.~\cite{yang2023how} revealed that VGG19 excels in predicting human correlation judgments on scatterplots and exhibits generalization ability across different designs.

In this work, we follow the above approach to evaluate the generalization performance of CNNs and humans on relational reasoning tasks involving bar charts. To the best of our knowledge, we construct the first generalization benchmark dataset for graphical perception tasks, which consists of a large set of synthesized standard visualizations and their element-wise perturbed counterparts with exposed and programmable parameters. We analyze the effect of perturbation operations on CNNs and compare the results with human judgments under the same conditions.

\section{Background}\label{sec:background}

In this section, we briefly review two fundamental experiments that evaluate the graphical perception of humans and CNNs.
We then revisit their hypotheses, experimental setups, and derived conclusions and 
report three major limitations that motivated our work.

\subsection{Cleveland and McGill's Experiments} \label{sec:cleveland-mcgill}
To study how humans decode graphs, Cleveland and McGill~\cite{cleveland1984graphical} summarized ten elementary graphical encodings from various commonly-used plots and charts: \emph{position along a common scale}, \emph{positions along non-aligned scales}, \emph{length}, \emph{direction}, \emph{angle}, \emph{area}, \emph{volume}, \emph{curvature}, \emph{shading}, and \emph{color saturation}. To evaluate the effectiveness of these visual encodings, they conducted two experiments using bar and pie charts. Here, we focus on the \emph{position-length} experiment, which estimates bar length ratios.

\vspace{1.5mm}
\noindent\textbf{Position-length experiment.}
Subjects were presented with a set of grouped and stacked bar charts with five different types, where two bars or bar segments were marked with black dots on each chart. The task was to estimate the ratio between the two marked bars, \ie the percentage of the smaller value to the larger value. Subjects could make judgments based on the \emph{position along a common scale} or the \emph{length} of the bars.

\vspace{1.5mm}
\noindent\textbf{Measurements.}
The perception accuracy was  measured by the midmeans of log absolute errors (MLAE) for each experimental unit in two experiments: 
\begin{equation} \label{mlae}
    	\textnormal{MLAE}(\hat{\theta}) = \frac{2}{n} \sum\nolimits_{i=0.25 \cdot n+1}^{0.75\cdot n}
log_2( | {\hat{\theta}_i} - {\theta}_i | + 0.125),
\end{equation}
where
$\hat{\theta} = \{\hat{\theta}_{1},\cdots,\hat{\theta}_{n}\}$ is a set of ordered estimations,
$\hat{\theta}_i$ is a predicted percent given by a subject, and $\theta_i$ is the corresponding true percent.

\vspace{1.5mm}
\noindent\textbf{Findings.}
Cleveland and McGill found that adjacent bars score the best, closely followed by the separated bars and horizontally aligned stacked bars, while unaligned stacked bars and vertically aligned bars are the worst.
The quantitative results show that the position judgments were 1.4 to 2.5 times more accurate than length and 1.96 times more accurate than angle.
The authors attributed the differences to different visual estimation strategies in humans, where aligned bars involve the judgment of positions along a common scale and unaligned bars involve length judgments.

\subsection{Haehn et al.'s Evaluation of CNNs}
To investigate if CNNs meet or outperform humans on graphical perception,
Haehn et al.~\cite{haehn2019evaluating} conducted five perceptual experiments using four off-the-shelf neural networks: a multilayer perceptron (MLP), LeNet~\cite{lecun1998gradient}, VGG19~\cite{simonyan2014very}, and Xception~\cite{chollet2017xception}, with increasing sophistication in the architectures.
They reproduced Cleveland and McGill's \emph{position-length} and \emph{position-angle} experiments and added three new experiments: (i) Cleveland and McGill's \emph{elementary perceptual tasks} to study if CNNs can extract quantities from basic visual marks, (ii) \emph{bars and framed rectangles} experiment to test if visual cues can help CNNs perform more accurately, and (iii) \emph{Weber's law point cloud} experiment to check whether CNNs and humans have a similar mechanism on perceiving differences. In this work, we mainly focus on the \emph{position-length} experiment.

\vspace{1.5mm}
\noindent\textbf{Measurements.}
In all experiments, Haehn et al. simplified the charts from Cleveland and McGill's study by removing the axes and text. They presented the CNNs with sketch-style black-and-white images and tested their predictions on charts with the same parameters. Then they compared the results with human performance data collected by themselves using the same MLAE metric as Cleveland and McGill.

\vspace{1.5mm}
\noindent\textbf{Findings.}
They found that CNNs performed worse than humans on the position-length experiment and had similar accuracy across different bar chart designs.
Overall, CNNs may exceed or perform similarly to humans for some experiments but cannot complete tasks such as the ones in the \emph{position-length} experiment, which requires estimating ratios between the lengths of the bars of interest. Therefore, Haehn et al. concluded that ``CNNs are not currently a good model for human graphical perception.''

\subsection{Study Limitations}
We identified the following three limitations of these studies:
\begin{enumerate}
	\item[\textbf{L.1}] Haehn et al. explored four network architectures but did not study the influence of hyper-parameters such as different optimization solvers and learning rates. They also did not investigate more advanced CNN model architectures such as ResNet~\cite{he2016deep}).
	
	\item[\textbf{L.2}] The charts that were used in the above studies are oversimplified. For example, they do not contain many indispensable elements, \eg axes, labels, or titles. The graphical perception performance on these charts may not reflect the actual capability of CNNs for standard visualizations, let alone the ones used in real applications.
	
	\item[\textbf{L.3}] Although humans are known to generalize effectively on common visual relational reasoning tasks, it remains unclear how CNNs perform when there is a discrepancy between the test and training data, as Haehn et al. only evaluated the generalization ability of CNNs when the test visualizations had similar visual encodings as the training ones.
\end{enumerate}

\section{Revisiting Haehn et al.'s Experiment}\label{sec:model-selection}

In this section, we replicate Haehn et al.'s evaluation of graphical perception with CNNs to
address limitation \textbf{L.1} and then, while still using the black-and-white chart images, we gradually expand the parameter space for data generation of perceptual tasks in four steps, resulting in an increased task complexity. We measure the performance of each step to see if CNNs can still accomplish the ratio estimation task.

\subsection{Replicating Haehn et al.'s Experiment}
\label{sec:controlled-settings}
Following Cleveland and McGill's task design and Haehn et al.'s experimental design, we reproduce the \emph{position-length} experiments using different network architectures and hyper-parameters.

In Cleveland and McGill's \emph{position-length} experiment, the authors used controlled parameters for the generation of the stimuli. The \emph{length values} involved in subjects' estimations (the marked target bars) were set as 10 real numbers
equally spaced on the log scale:
\begin{align}
	\label{equ:si-10}
	s_{i}=10^{1 + (i-1) / 12}, \quad i=1, \cdots, 10.
\end{align}
They also fixed the \emph{indices} of the target bars, for example, the second and third ones in the left group of the bar charts.
Furthermore, the \emph{pixel dots} (a one-pixel dot for each bar) were all located in a fixed position inside the target bars.

Haehn et al.~\cite{haehn2019evaluating} used the same parameters when conducting the experiments with CNNs.
In addition, they randomly sampled 
 length values of non-target bars from an interval $[10,93]$, guaranteeing that all the local patterns in the image could be detected by the largest convolutional filter.
They synthesized 100,000 $100 \times 100$ sized images containing sketch-styled bars in black and white.
To avoid the CNN models simply memorize the data, they added 5\% normally-distributed noise with $\textnormal{mean} = 0$ and $\textnormal{variance} = 1$ to each image.

\vspace{1.5mm}
\noindent\textbf{CNN Architectures.}
The network architecture is crucial for the performance of CNNs in various applications. Haehn et al. used VGG~\cite{simonyan2014very}, a simple and effective CNN architecture that supports up to 19 layers.
Since their result shows that VGG19 achieves the best graphical perception performance among other networks, we use VGG19 as one of the baselines in our experiments.
To investigate if a better performance can be achieved by more recent CNNs, we include a ResNet152~\cite{he2016deep} in our experiments, which is 8 times larger than VGG19 and deeper than the Xception126~\cite{chollet2017xception} network also used by~\cite{haehn2019evaluating}.
We also test multi-layer perceptron (MLP), AlexNet~\cite{kirzhevsky2012imagenet}, LeNet~\cite{lecun1998gradient}, DenseNet~\cite{densenet2017}, EfficientNet~\cite{efficientnet2019}, and a Relation Network~\cite{relationnetwork2017}.

\begin{table}
	\centering
	\caption{Performance of VGG19 and ResNet152 with different hyper-parameters in type 1 of the position-length experiment, where the cell with \xmark{} indicates that the corresponding parameter is invalid, and CI indicates confidence interval. Note that the human estimated MLAE value in this task reported by previous studies~\cite{haehn2019evaluating} is 1.4.
}
	\label{tab:networks-hyperparameters}
	\setlength{\tabcolsep}{3pt}
	\ra{1.2}
 \resizebox{\linewidth}{!}{
		\begin{tabular}{ccccccr}
			\toprule
				\multirow{2}{*}{\textbf{Network}} & \multirow{2}{*}{\textbf{Optimizer}} & \multirow{2}{*}{\textbf{\shortstack[C]{Learning\\ rate}}} & \multirow{2}{*}{\textbf{Momentum}} & \multirow{2}{*}{\textbf{\shortstack[C]{Weight\\ decay}}} & \multicolumn{2}{c}{\multirow{2}{*}{\textbf{MLAE\quad CI}}}   \\
			&                                     &                                                                                   &                                    &          &  &                                         \\ \midrule
			\multirow{8}{*}{VGG19}            & \multirow{4}{*}{SGDM}               & \multirow{2}{*}{High}                                                             & Classic                            & \xmark                                                            & 1.43& {0.05}                                          \\
			&                                     &                                                                                   & Nesterov                           & \xmark                                                             & 1.10& {0.22}                                       \\ \cline{3-7}
			&                                     & \multirow{2}{*}{Low}                                                              & Classic                            & \xmark                                                            & 1.41& {0.15}                                          \\
			&                                     &                                                                                   & Nesterov                           & \xmark                                                            & 1.04& {0.27}                                       \\ \cline{2-7}
			& \multirow{4}{*}{AdamW}               & \multirow{2}{*}{High}                                                             & \xmark              & High                                                                             & -2.32& {0.11}                                          \\
			&                                     &                                                                                   & \xmark              & Low                                                                              & \textbf{-2.41}& {0.10}                                           \\ \cline{3-7}
			&                                     & \multirow{2}{*}{Low}                                                              & \xmark              & High                                                                             & -1.89& {0.35}                                             \\
			&                                     &                                                                                   & \xmark              & Low                                                                              & -1.71& {0.25}                                             \\ \midrule
			\multirow{8}{*}{ResNet152}        & \multirow{4}{*}{SGDM}               & \multirow{2}{*}{High}                                                             & Classic                            & \xmark                                                            & -0.61& {0.24}                                             \\
			&                                     &                                                                                   & Nesterov                           & \xmark                                                            & -0.60& {0.21}                                            \\ \cline{3-7}
			&                                     & \multirow{2}{*}{Low}                                                              & Classic                            & \xmark                                                            & 0.72& {0.24}                                            \\
			&                                     &                                                                                   & Nesterov                           & \xmark                                                            & 0.70& {0.25}                                           \\ \cline{2-7}
			& \multirow{4}{*}{AdamW}               & \multirow{2}{*}{High}                                                             & \xmark              & High                                                                             & \textbf{-2.76}& {0.10}                                           \\
			&                                     &                                                                                   & \xmark              & Low                                                                              & -2.42& {0.17}                                          \\ \cline{3-7}
			&                                     & \multirow{2}{*}{Low}                                                              & \xmark              & High                                                                             & -2.71& {0.10}                                            \\
			&                                     &                                                                                   & \xmark              & Low                                                                               & -2.53& {0.24}                                           \\ \bottomrule
		\end{tabular}%
}
\vspace{-3mm}
\end{table}

\vspace{1.5mm}
\noindent\textbf{Optimization Solvers.}
Optimizers also play an important role in the training process of CNNs and can have a big influence on their performance. Stochastic gradient descent (SGD) is a classic iterative method for optimizing deep neural networks. 
Haehn et al. use SGD with momentum (SGDM)~\cite{qian1999momentum} to improve the classic SGD with the advantages of faster convergence and fewer oscillations during training. 
More recently, there are more powerful and efficient optimization solvers for training deep neural networks. Adaptive moment estimation (Adam)~\cite{kingma2014adam} is a representative example, which achieves better performance than SGDM.
Loshchilov and Hutter~\cite{loshchilov2017decoupled} proposed an improved version of Adam called AdamW, which shows better training curves and generalization ability. In this work, we explore the different performances of SGDM and AdamW.

\vspace{1.5mm}
\noindent\textbf{Training Objective.}
Given the objective of predicting continuous bar length ratios, we train regression models as in previous work~\cite{haehn2019evaluating}.
Although MLAE (Equation~\ref{mlae}) is suitable for perception accuracy, using it as a log-scale loss during network training leads to biased gradients for different prediction error scales. Thus, similar to Haehn et al.~\cite{haehn2019evaluating}, we train our models using the mean squared error (MSE): 
\begin{align}
\textnormal{MSE}(\hat{\theta})=\frac{1}{n} \sum\nolimits_{i=1}^{n}\left(\hat{\theta}_{i} - \theta_{i}\right)^{2}. \nonumber
\end{align}
Following the suggestion of Haehn et al., we normalize both $\hat{\theta}_{i}$ and  $\theta_{i}$ to the range $[0,1]$. When MLAE is -3, MSE is around 0; when MLAE is 0, MSE is around $0.76$. A negative MLAE score of around -3 indicates high estimation accuracy, and a high positive score indicates poor accuracy.

\vspace{1.5mm}
\noindent\textbf{Hyper-Parameters Tuning.} \label{sec:model-selection-training}
Another issue for achieving optimal model performance is the selection of hyper-parameters during CNN training, such as learning rate, momentum type, weight decay, batch size, and epoch number.
We follow the suggestions by recent guidelines~\cite{diaz2017effective,hinz2018speeding,cnn_guide} for tuning hyperparameters. Table~\ref{tab:networks-hyperparameters} shows our chosen values for \emph{learning rate} and a parameter for each of the two main \emph{optimizers}: \emph{momentum type} for SGDM and \emph{weight decay} for AdamW.
To find proper values for learning rate and weight decay, we adopt a coarse-to-fine strategy that first searches for coarse ranges of hyperparameter values using a small number (100) of epochs and a small training set (1/3 of the entire dataset) and gradually narrows the range down through binary search.
To alleviate falling into a local minimum,  we search for the optimal values from two different ranges: one with large values and the other using small values. Hence, we set two learning rates for each combination of network architectures and optimizers to inspect if higher or lower values contribute to better results.

For SGDM, we test two commonly used types of momentum, classic and Nesterov momentum.
For AdamW, we examine the effect of weight decay and set two values as high and low, respectively.
Based on these parameters, we follow the previous guidelines that train the networks for 500 epochs with a mini-batch size of 16 and 32 for each task and terminate early if the validation loss stays unchanged for ten epochs.

Since Haehn et al.~\cite{haehn2019evaluating} reported poor performance of CNN models that were pre-trained on ImageNet~\cite{kirzhevsky2012imagenet}, we do not perform any pre-training and train the models from scratch.
For all the experiments, we employed 5-fold cross-validation for each model. We split the dataset into five equal-sized subsets and used four subsets for training and one for validation iteratively. This process was repeated five times, ensuring each subset was selected as the validation set exactly once.
Our models are implemented in Pytorch and trained on an NVIDIA RTX 3090 GPU with mini-batch size 16 for 500 epochs.
All networks are initialized with random parameters.

\begin{figure}[t]
	\centering
	\includegraphics[width=\columnwidth]{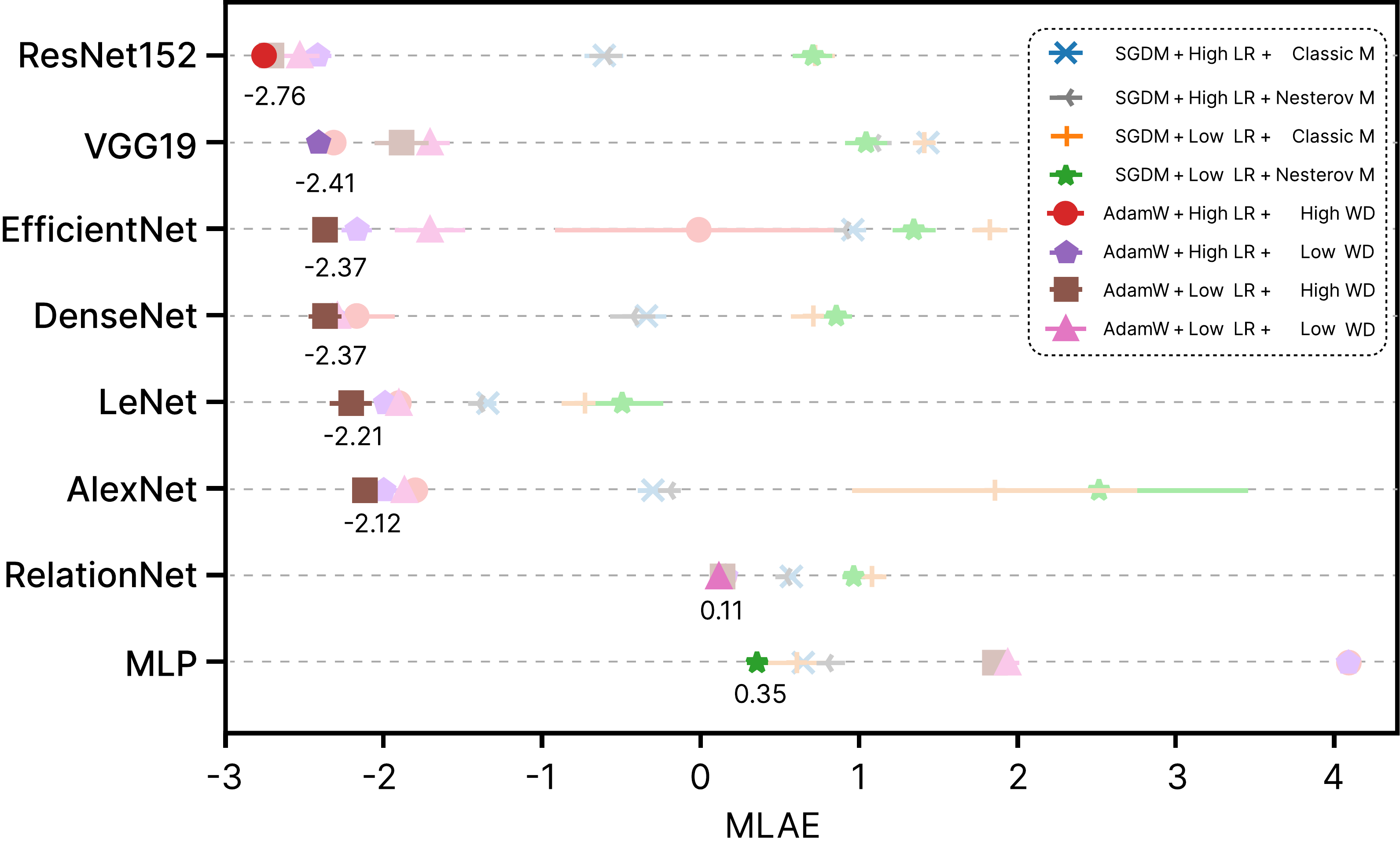}
	\caption{Performance comparison of eight network architectures trained with eight sets of hyper-parameters in type 1 of the position-length experiment. The best-trained model for each kind of network is highlighted and labeled with the corresponding MLAE value, while others are shaded. 
 }
	\label{fig:networks-performance}
    \vspace{-3mm}
\end{figure}

\vspace{1.5mm}
\noindent\textbf{Results.}
All models outperform VGG19 (MLAE=3.51) as reported by~\cite{haehn2019evaluating}, even the simple MLP (MLAE=0.35).
Given this unexpected result, we first reviewed the training process by comparing the training and validation loss curves~\cite{brownlee2018better} to rule out issues such as overfitting.
The curves are provided in the supplemental material, where the validation loss consistently decreased alongside the training loss, confirming that the training process was conducted properly.

Across the eight tested network architectures, VGG19 and ResNet152 perform significantly better than other networks as shown in Fig.~\ref{fig:networks-performance}. 
To analyze the influence of hyper-parameters, we report the detailed results of these two networks for the position-length type-1 experiments in Table~\ref{tab:networks-hyperparameters}, as similar patterns are observed in the remaining models
Comprehensive results are provided in the supplementary material.
Haehn et al.~\cite{haehn2019evaluating} reported the lowest MLAE=3.51 achieved by VGG19 compared to humans (MLAE=1.4) for this task, which led to their conclusion that CNNs \emph{cannot} complete this task.
In contrast, we find that \textbf{VGG19 and ResNet152 trained with carefully selected hyper-parameters \emph{exceed} human performance}, as shown in Table~\ref{tab:networks-hyperparameters}.

Examining the influence of each factor,
we find that AdamW enables VGG19 and ResNet152 to produce the best results, with a higher learning rate improving network performance in most cases.
In contrast, the influence of momentum and weight decay are both unclear as the MLAE grows higher or lower for different conditions. The batch size has a weak influence on model training since smaller (16) and bigger (32) sizes result in similar MLAEs.
Throughout all combinations of networks and hyper-parameters, ResNet152 trained with AdamW with a high learning rate and weight decay performs best for the \emph{position-length} experiments. In particular, the MLAE value for the experiment is $-2.76$, which is highly accurate and much better than average human performance (1.4). 
Therefore, we perform further evaluations using this configuration of ResNet152 with AdamW.

\subsection{Expanding the Data Generation Parameter Space}

\label{sec:realistic-settings}

In this experiment, we use the same black-and-white chart images as in Section~\ref{sec:controlled-settings}, but gradually expand the parameter space for data generation 
in three steps, resulting in increased task complexity. We measure the performance of each step.

\begin{figure}[!b]
	\centering
	\includegraphics[width=0.9\columnwidth]{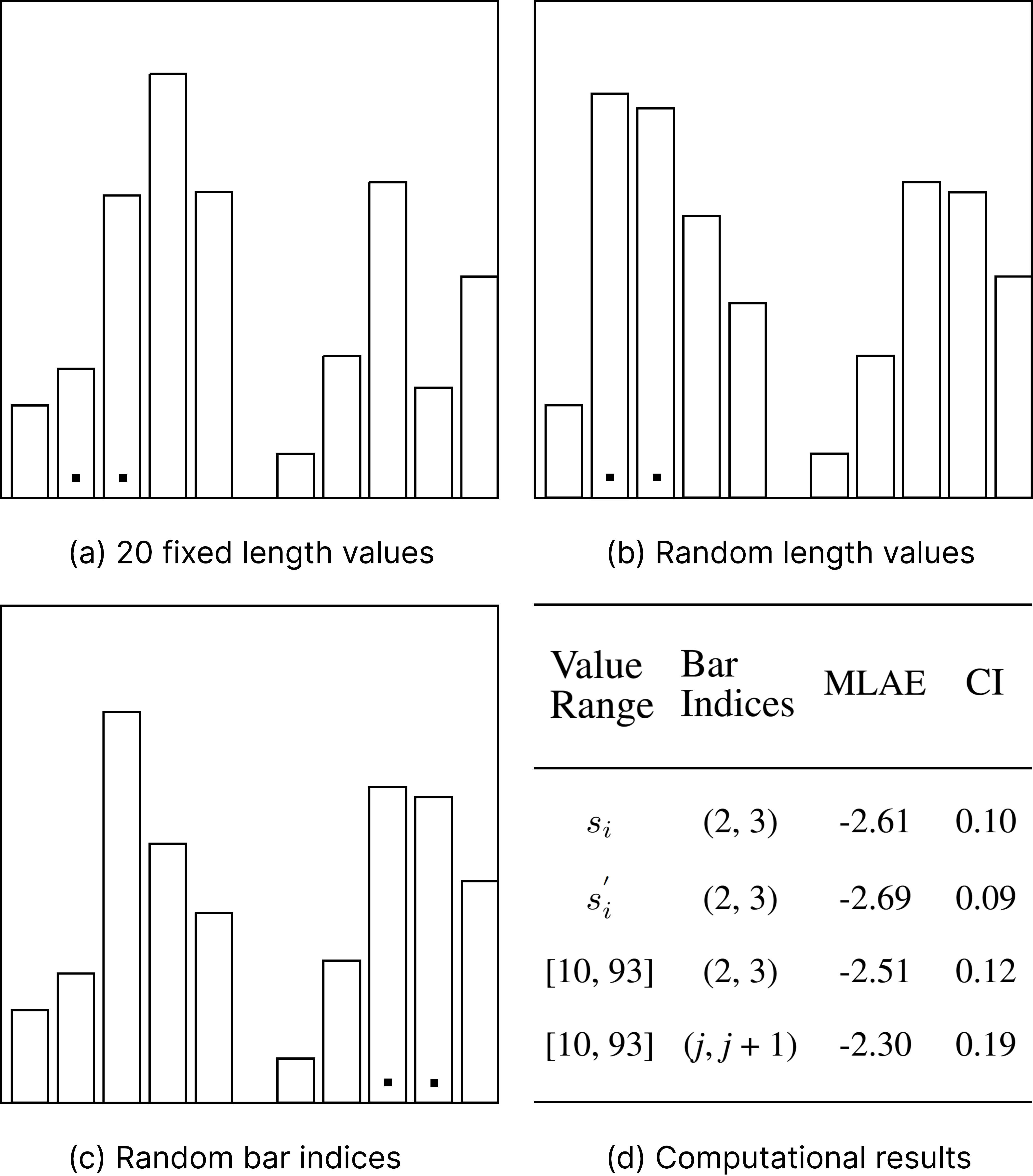}
	\caption{Sample images of expanding the parametric space in three steps: (a) from 10 fixed length values to 20 fixed values, (b) random length values,  and (c) from fixed indices of target bars to random indices; (d) control settings and MLAE values and CIs of three extended steps in (a-c)}.
	\label{fig:realistic-settings}
\end{figure}

\vspace{1.5mm}
\noindent\textbf{Experimental Setup.}
As shown in Fig.~\ref{fig:realistic-settings}, the appearance of the bars to be estimated is mainly decided by two factors: the number of possible bar heights, and indices of the target bars. Here, we investigate how CNNs perform when expanding the value range of each factor using the following steps:

\begin{itemize}
\item[(i)] We expand the range of target bar lengths from 10 to 20 fixed numbers by modifying Equation~\ref{equ:si-10}:
\begin{align}
	s_{i}^{'} = 10^{1+ (i-1) / 24}, \quad i=1, \cdots, 20. \nonumber
\end{align}
\item[(ii)] We further expand the length range of all bars to random numbers within the interval $[10,93]$. This range is chosen as in~\cite{haehn2019evaluating} to ensure that all content in a $100\times 100$ image can be seen by large convolutional filters. This also allows for testing on out-of-distribution lengths in the future (see Section~\ref{sec:generalization-performance}).
\item[(iii)] Instead of fixing the target bars (indicated by black dots) to the second and third ones, we define them as two arbitrary adjacent bars $(j,j+1)$, where $j$ ranges from 1 to 9.

\end{itemize}

The dataset generated (consisting of 60,000 samples for training, 20,000 for validation, and 20,000 for testing) in each step is fed into the network for model training and evaluation. Later, we test the model with unseen stimuli.
We use ResNet152 with the AdamW optimizer and keep all the other hyper-parameter settings the same as in Section~\ref{sec:model-selection-training}.

\vspace{1.5mm}
\noindent\textbf{Results.}
The test MLAEs of the CNN models on our expanded parameter space are shown in Fig.~\ref{fig:realistic-settings}(d).
A reasonable hypothesis is that the model performance will drop as the datasets and tasks become more complicated.
This is confirmed by the general trend of our experimental results.
Despite a performance decrease, the MLAE value of our CNN models is -2.30, already reaching an extremely high estimation accuracy that is significantly better than that of humans (average MLAE=1.4).
Hence, \textbf{CNNs perform well in the expanded data generation parameter space and achieve comparable performance as in the previously highly controlled settings.}
However, even in such a setting, the stimuli are still far simpler with no essential graphical elements adjunct to the visualizations like axes and texts or using color. 

\section{Generalization of CNNs and Humans on Standard Visualizations}
In this section, we explore how well CNNs and humans perform on standard visualizations.
To address \textbf{L.2} and \textbf{L.3}, we first create \textbf{GRAPE} (GRAphical PErception tasks), a synthetic dataset that contains 766,000 (500,000 for training and 266,000 for testing) images of standard visualizations generated by Vega-Lite~\cite{satyanarayan2016vega}.
Next, we train a CNN model for each task and test how well they perform on unseen visualizations. We also conduct a user study to measure human performance. 

Since Haehn et al.~\cite{haehn2019evaluating} only evaluate CNNs with test data that has similar visual encodings as the training data, their results do not allow to reveal how well CNNs handle charts with slight variations in visual encodings (\textbf{L.3}).
Hence, we create two kinds of test stimuli based on GRAPE: one sharing similar visual encodings with the training data, and the other with variations in the visual encodings. We refer to these two settings as independent and identically distributed (IID) and out-of-distribution (OOD) test data, respectively~\cite{geirhos2018generalisation,geirhos2021partial}.

\subsection{GRAPE Dataset}
\label{sec:standard-vis}

In contrast to the simple black-and-white charts in the previous studies~\cite{cleveland1984graphical, haehn2019evaluating}, the charts used in our next set of experiments are standard visualizations that were generated by Vega-Lite~\cite{satyanarayan2016vega}.
Here, we describe the construction of our dataset of standard visualizations, \emph{GRAPE}, and the methodology we followed to create training and test stimuli for CNNs.

\vspace{1.5mm}
\noindent\textbf{Parametric Space.}
Vega-Lite uses the grammar of graphics~\cite{wilkinson2012grammar} with a set of parameters that control the visual appearances of independent chart elements. We programmatically vary commonly used parameters for bar charts, such as bar width, stroke width, title properties (position, font, and size), and colors (background color, bar color, and stroke color).
Our parametric model also supports other parameters like tick number, axis label, center position, etc.

\begin{figure*}[!t]
	\centering
	\includegraphics[width=0.9\textwidth]{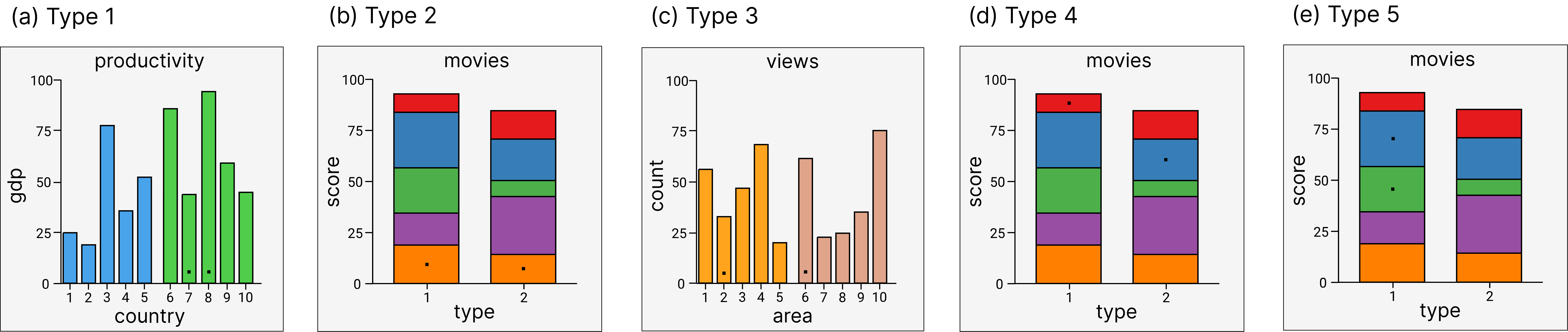}
	\caption{Our five types of stimuli used in the \emph{position-length} experiment, where each bar chart includes colorized bars, axes, tick labels, and titles.}
	\label{fig:ours-position-length-exhibition}
    \vspace{-3mm}
\end{figure*}
Since color is an essential component of standard visualizations, we design a color mapping scheme that satisfies two requirements: adjacent colors have visually noticeable differences, and all colors are nameable to facilitate communication.
To do so, we take the 140 named colors from D3~\cite{bostock2011d3}, order them by their perceived luminance, and divide them into four non-overlapping color sets.
We assign the brightest (set 1) and darkest (set 4) colors to the background and stroke, respectively.
Set 3 is used for the bar fill color, while set 2 is used to generate completely unseen color distributions for testing our CNN models.
To ensure sufficient color discriminability, we make sure that the luminance difference between the colors of any pair of visual elements (e.g., bars, strokes, and background) in one chart is at least 5 in CIE LAB color space.

With the above parameter settings, we generate a set of chart images of size $150\times150$ pixels. The reason that we increased the image size, compared to the $100\times 100$ pixels used by Haehn et al.~\cite{haehn2019evaluating}, is to provide enough space to place the title, axis, and labels.
Fig.~\ref{fig:ours-position-length-exhibition} shows the resulting five types of bar charts for the \emph{position-length} experiments. Like Haehn et al., we add 5\% random noise (normally distributed between -0.025 to 0.025) to each pixel to prevent the networks from memorizing each image.

\vspace{1.5mm}
\noindent\textbf{Training Stimuli.}
We increase the task complexity of the \emph{position-length} experiment setting of Haehn et al.~\cite{haehn2019evaluating} by adding two aspects. 
(i) One extension is to expand the two parameters (length range and indices of target bars) defined in Section~\ref{sec:realistic-settings}.
As shown in the last row of Fig.~\ref{fig:realistic-settings}(d), the value range of each bar is extended to $[10,93]$. For stacked bar charts, we further normalize the total values within each stack.
For the indices of the target bars, we follow Cleveland and McGill, who used any two adjacent bars for types 1 and 5, random bars from different groups/stacks for types 3 and 4, and the bottom two bar segments for type 2 (Fig.~\ref{fig:ours-position-length-exhibition}).
(ii) The other extension is to sample the ten bar values from more data distributions with different patterns (\eg long tail, peak, and sine wave) in addition to the random distribution.
All stimuli are colored instead of the previous over-simplified black-and-white ones. For each task, we generate $60,000$ images for training, $20,000$ for validation, and $20,000$ for testing.

\begin{figure*}[!t]
	\centering
	\includegraphics[width=0.9\textwidth]{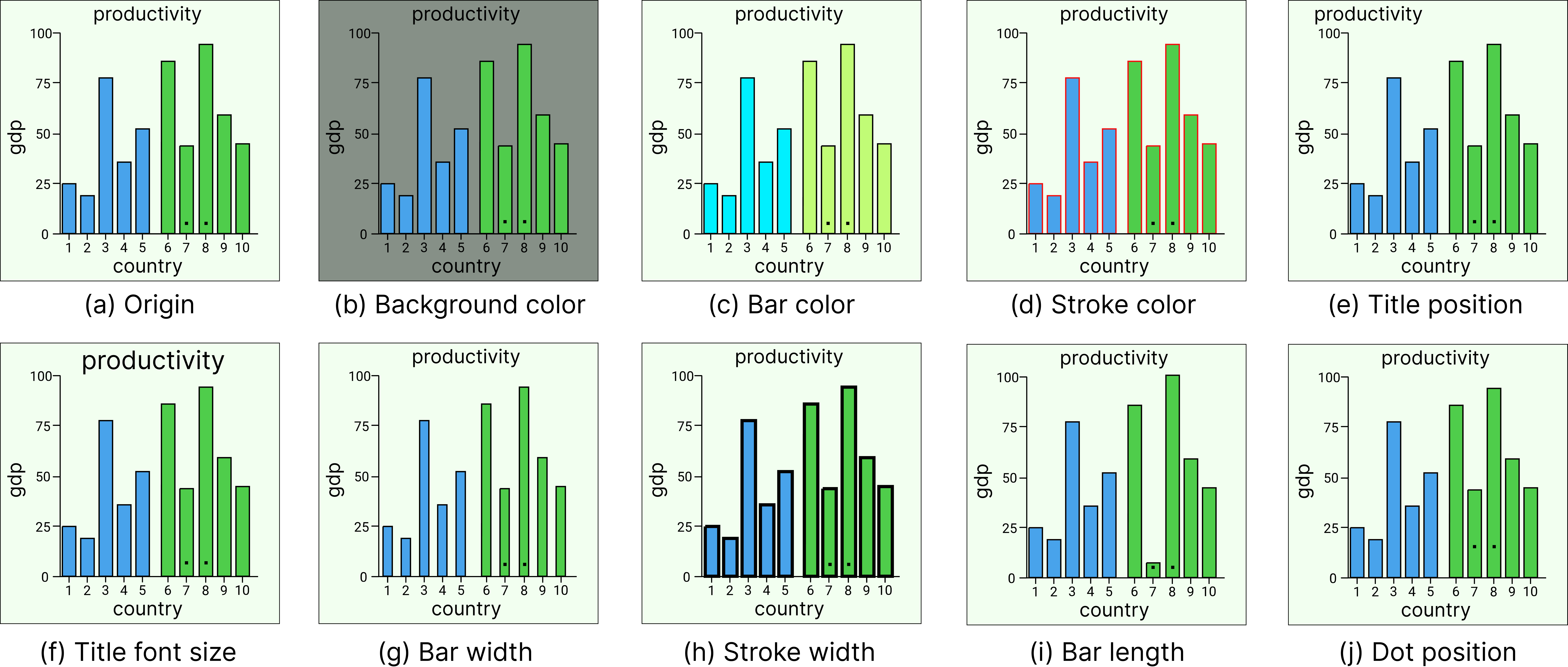}
	\caption{An exemplar training stimulus (a) of type-1 and the test stimulus (b-j) generated by perturbing eight different parameters of one bar chart of type-1 at one specific level.} 
\label{fig:parameter-example}
\vspace{-3mm}
\end{figure*}

\vspace{1.5mm}
\noindent\textbf{Test Stimuli.} \label{sec:robustness}
Our test dataset comprises two types of stimuli procedurally generated for IID and OOD test respectively. For the IID test, the charts are generated using the same parameters as in the training stimuli.
To investigate how CNNs perform on unseen visualizations, especially ones with variations in parametric space,  we generate new test stimuli for the OOD test by perturbing the visual encoding parameters.
We choose to manipulate 9 parameters that can be divided into three categories: global chart properties (title position, title font size, and background color), \emph{task-unrelated} mark properties (bar width, bar color, stroke width, and stroke color), and \emph{task-related} mark properties (bar length and dot position). \hl{Fig.~\ref{fig:parameter-example} depicts one exemplary perturbation for each of the nine parameters of the type 1 bar chart.}

For each parameter, except bar length, we gradually increase and decrease the corresponding value in the training stimuli, such as luminance for color-related parameters. For example, chart titles are center-aligned in the training set, whereas in the test stimuli they are gradually moved to the left or right by 15\% of the canvas width each time.
Since the color sets assigned to the background and stroke consist of the brightest and darkest colors, respectively, we only decrease the luminance of the background colors and increase the luminance of the stroke colors. If the perturbed value of one parameter ends up out of the valid range, we discard the corresponding stimuli. The bar length is determined by the encoded data value. Since the range of data values in the training stimuli is $[10,93]$, we generate the bar lengths for the test stimuli in the value range $[1,9] \cup [94,100]$.

By default, the perturbations of each parameter have six levels, as shown on the x-axis of Fig.~\ref{fig:parameter-effect}. For example, the title position is moved by 15\% of the canvas width three times from the center to both the left and right direction, respectively. To ensure that all marks are within the canvas and do not overlap with each other, we only set four levels of perturbation for bar width on types 1 and 3, and two levels for stroke width on all types of bar charts.
As for dot position, we only move it along the horizontal direction for stacked bar charts (type 2,4,5) and the vertical direction for grouped bar charts (type 1,3).
For each level of the perturbation, we use the same method described earlier in this section to generate 1,000 chart images. 
Taking all types and all levels of perturbed charts together results in a total number of $266,000$ test stimuli.

\subsection{Generalization Performance of CNNs}
\label{sec:generalization-performance}

\begin{figure*}[h]
	\centering
	\includegraphics[width=\linewidth]{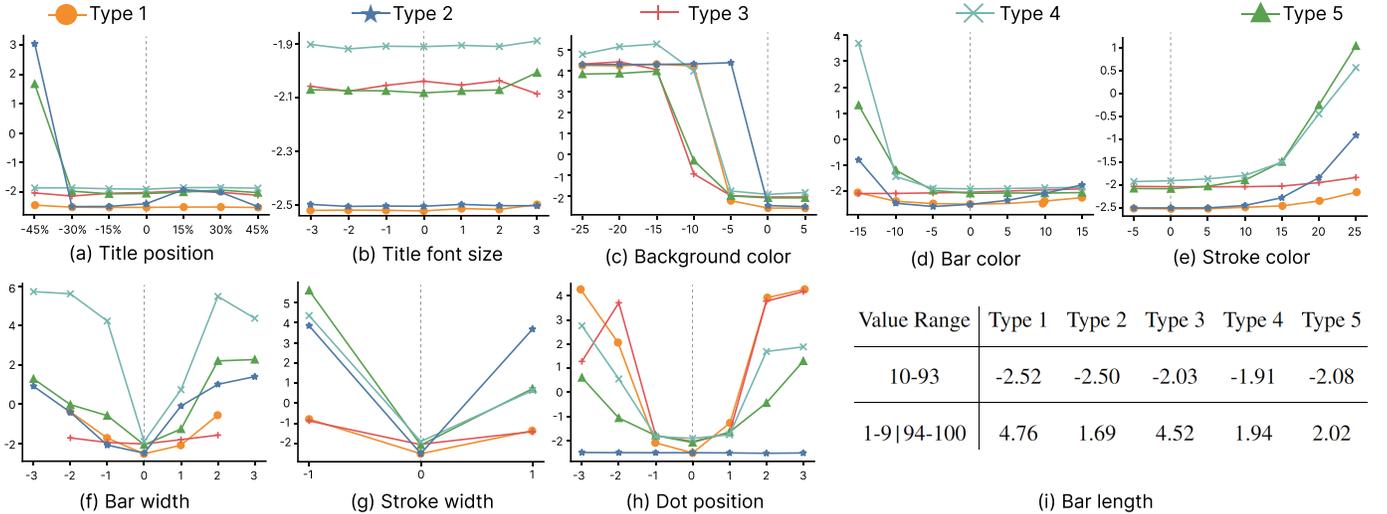}
	\caption{The mean MLAE values produced by CNNs on performing generalization tests of eight parameters on five types of bar charts. (a-h) Each curve shows how the MLAE values change by increasing or decreasing the corresponding parameter values. The dotted lines indicate the MLAE value computed for the stimuli with non-perturbed parameters. (i) Estimated MLAE values for bar charts with bar lengths encoded by different value ranges.}
	\label{fig:parameter-effect}
 \vspace{-3mm}
\end{figure*}

With our new GRAPE dataset, we re-train the different model configurations described in Section~\ref{sec:model-selection} for each type of chart to test the IID generalization on standard visualizations. Then, we apply the trained CNN models to each set of perturbed test visualizations to benchmark their OOD generalization. Fig.~\ref{fig:parameter-effect} shows the results.

\vspace{1.5mm}
\noindent\textbf{IID Test.}
The MLAE values corresponding to the zero perturbation value on the x-axis of Fig.~\ref{fig:parameter-effect}(a-h)
and the $[10,93]$ interval on the value range column of Fig.~\ref{fig:parameter-effect}(i)
are the average MLAE values of CNN estimations on standard visualizations with our expanded settings.
Even though more complicated datasets are fed into the CNNs, our results are significantly better than Haehn et al.'s~\cite{haehn2019evaluating}.
The test MLAE values for the five types of bar charts are all between $-2.52$ and $-1.91$, whereas even the best performance reported by Haehn et al. for the same tasks is above 3.

Compared with the results in Section~\ref{sec:model-selection}, where CNNs were trained with black-and-white images under highly controlled settings, the overall MLAEs on standard visualizations only dropped less than .5 for all types of tasks. These average MLAE values (all negative) are also significantly lower than those of humans reported by Haehn et al.~\cite{haehn2019evaluating}, which are 1.4 on the over-simplified stimuli. 
In summary, our results show that \textbf{CNNs can achieve extremely high IID generalization performance on graphical perception tasks.} They produce unbiased estimations for all types of bar charts when both training and test data satisfy the IID property.

\vspace{1.5mm}
\noindent\textbf{OOD Test.}
The curves in Fig.~\ref{fig:parameter-effect} show how MLAE changes as a function of the perturbation levels of each parameter. We see that MLAE values are almost the same for different perturbations of the title position (Fig.~\ref{fig:parameter-effect} (a)) and title font size (Fig.~\ref{fig:parameter-effect} (b)), except for type 2 and 5 on the -45\% perturbation of title position. For the three color parameters (Fig.~\ref{fig:parameter-effect} (c-e)), CNNs are least robust to the background color as the MLAE values of CNNs' estimation on all types of charts change to around 4.0 when the luminance value decreases by 15. This might be reasonable since the background color affects a larger area and can significantly change the appearance of the whole chart compared to the colors of other visual marks.

While the MLAE values do not change much for minor perturbations of color and title parameters, we find that a small positive or negative change to some parameters like bar width, stroke width, and dot position causes significant changes, forming the symmetric curves in Fig.~\ref{fig:parameter-effect}(f-h).
For example, increasing or decreasing just one pixel of stroke width results in a change from $-2.5$ to $4.0$ of the MLAE value on type-2 bar charts.
As for the bar length encoded by the value ranges $[1,9] \cup [94,100]$, the MLAE values of types 1 and 3 are even larger than 4.0, whereas for types 2 and 4, the values are less than 2.0.
These results indicate that the CNNs might not have modeled the ratio estimation task correctly, as many task-unrelated visual properties heavily impact its relational reasoning.

Among these five types of bar charts, only type 2 is robust to the perturbation of dot position, which is used to indicate the target bars (Fig.~\ref{fig:parameter-effect}(h)).
We speculate that the reason for this phenomenon is that the target bars of type 2 are restricted to the only two ones on the bottom of the chart, \ie the indices are actually fixed, which makes it easier for a CNN to model the task. In contrast, the MLAE values of type 2 and 5 all show significant changes when the title is moved to the leftmost position (Fig.~\ref{fig:parameter-effect}(a)). This is unexpected since the title position is unrelated to the target bars. For the stroke color and bar color, types 2, 4, and 5 are less robust to large perturbations than the other types. We speculate that the relatively large color complexity (i.e., more colors are placed closely) in the three stacked bar charts makes CNNs more sensitive to variations of stroke and bar colors.

In summary, we report two findings for OOD generalization: (i) \textbf{the relational reasoning ability of CNNs is heavily influenced by most visual parameters, regardless of whether they are related or unrelated to the target visual marks;} and
(ii) \textbf{the generalization performances of trained CNN models are almost equal on different types of bar charts in the IID setting, but vary significantly in the OOD setting.} Combining these observations, we draw the conclusion that despite their outstanding performance on ordinary IID tests, our used CNNs have not truly learned how to solve the ratio estimation task correctly.

\subsection{Comparison with Human Performance} \label{sec:user-study}

To further check whether the generalization performance of CNNs aligns with humans, we conduct a user study to perform the same tasks with human subjects in the IID and OOD settings
as described in Section~\ref{sec:generalization-performance} and then compare their results.
To obtain high-quality human data, we 
perform the ratio estimation task in a well-controlled lab environment with sufficient control over the physical workspace, display calibration, and observer attention.

\vspace{1.5mm}
\noindent\textbf{Measure.}
Before describing our study, we first define \emph{MLAE deviation} to quantitatively measure the influence of different perturbations on CNNs and humans:
\begin{align}
	\Delta \textnormal{MLAE}(\hat{\theta}) = \textnormal{MLAE}(\hat{\theta}_{ood}) - \textnormal{MLAE}(\hat{\theta}_{iid}),
\end{align}
where $\hat{\theta}_{iid}$ is an estimated value of the chart without perturbation, $\hat{\theta}_{ood}$ is the corresponding predicted value of the same target after a perturbing operation is performed.
A larger deviation means the perturbation has a greater influence.

\vspace{1.5mm}
\noindent\textbf{Experiment Design.}
In this experiment, each participant is asked to estimate length ratios of bar charts generated by applying different levels of perturbations on eight parameters. \hl{The purpose is to examine whether these perturbations affect human perception in a similar way as they do for CNNs. To this end,} there are two kinds of conditions: parameters and perturbation levels.
Note that we do not include the title font size, because its perturbation almost brings no change to the MLAE values of CNNs (Fig.~\ref{fig:parameter-effect}(b)). 

However, the number of all stimuli used for evaluating the generalization performance of the CNNs is too large for a human subject study. For example, in our automatic CNN tests, the background color has 7 levels. If we select 5 images for each level of all five types, we end up with $5\times7\times5 = 175$ bar charts, \hl{which would be excessively demanding for human subjects if presented directly.}

To address this issue, we introduce a \emph{two-phase} experimental scheme based on the previous finding that small perturbations do not affect human judgment significantly~\cite{geirhos2018generalisation}.
\hl{Generally, we first identify the visual parameters that have the greatest impact on human judgments and then evaluate how humans respond to different levels of perturbations applied to the identified parameters. If human subjects consistently perform well even as the most influential parameters are altered, we can infer that humans are robust to variations in visual parameters in relational reasoning tasks on bar charts.}

Specifically, in the first phase, we evaluate human performance on five types of bar charts with the largest perturbations to each of the visual parameters in Fig.~\ref{fig:parameter-example}. Here, we have eight conditions, where each condition corresponds to one visual parameter.
To further reduce the number of trials, we ask each participant to complete the tasks of all eight parameters for one type of chart and recruit the same number of different participants for each chart type.
Then, we identify the top three (type, parameter) pairs that affect human performance the most, and conduct progressive perturbing experiments on these pairs as in Section~\ref{sec:standard-vis} in the second phase. Since our goal is to investigate how different perturbation levels of the three most influential parameters affect human estimation accuracy, we ask each participant to complete the trials generated by all perturbation levels of all three pairs.
Thus, the first phase of our experiment is a \emph{between-subject} design where participants are divided into five groups and each group completes the trials of one type of bar chart, while the second phase is a \emph{within-subject} design to evaluate the effect of different perturbation operations.

\begin{figure*}[h]
	\centering
    \includegraphics[width=0.98\textwidth]{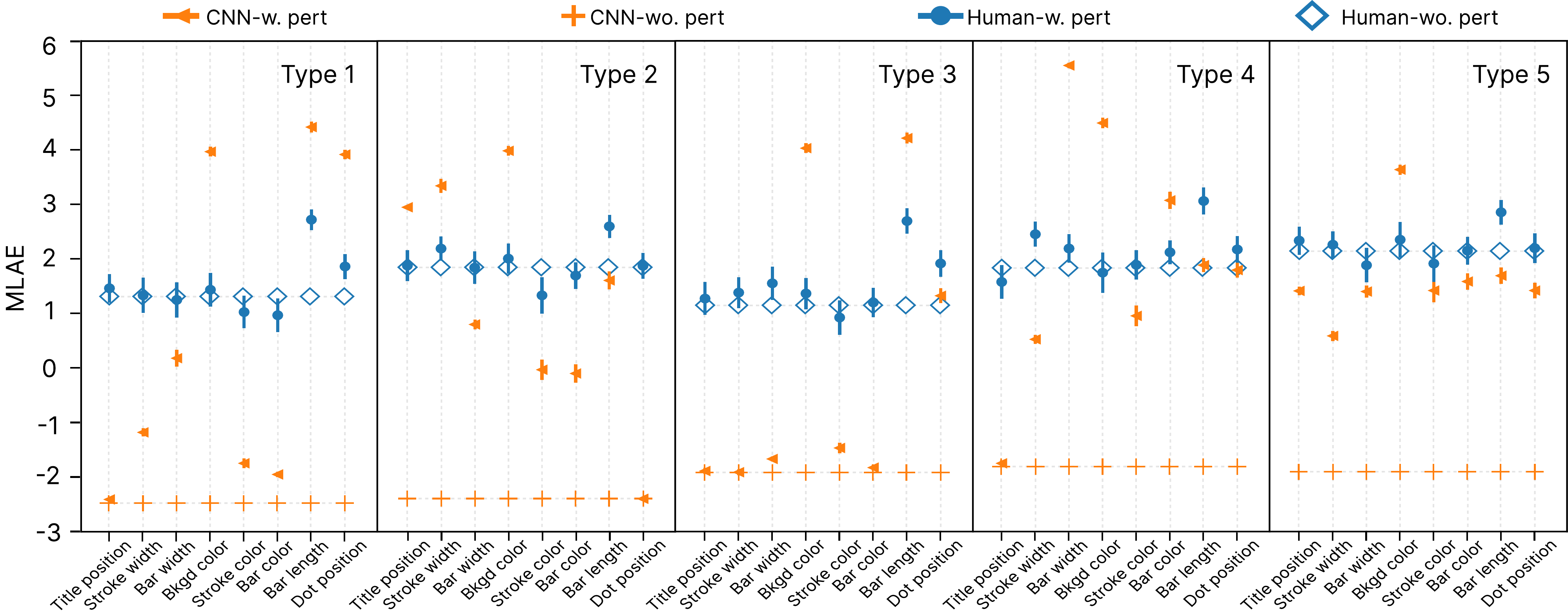}
	\caption{Comparison of bar chart tasks produced by humans and CNNs: mean MLAE and 95\% confidence interval values produced by humans and CNNs on five types of bar charts without and with the largest level perturbations on eight parameters.}
	\label{fig:human-cnn-comparison}
 \vspace{-3mm}
\end{figure*}
\noindent\textbf{Device.}
The study was run on a quad-core PC with a 23-inch color-calibrated screen monitor (120 Hz refresh rate) using a mouse as input and a $1920 \times 1080$ pixel display. All participants were seated with their eyes approximately 60 cm from the display in a quiet room.

\vspace{1.5mm}
\noindent\textbf{Hypotheses.}
As we saw, CNNs perform significantly better than humans for the over-simplified charts in Section~\ref{sec:model-selection}. Other studies have shown, however, that they are less robust than humans to image manipulations~\cite{geirhos2018generalisation}.
Thus, we formulate the following two hypotheses:
\begin{enumerate}
	\item[H1:] Human performance is inferior to CNNs in terms of estimation accuracy on standard visualizations; and
	\item[H2:] Human performance is more robust to perturbations of various visual parameters than that of CNNs.
\vspace{-2mm}
\end{enumerate}

\vspace{1.5mm}
\noindent\textbf{Participants.}
We recruited participants with color-normal vision from our local universities and communities.
Since bar charts are common visualizations, we follow the conclusion from Hall et al.~\cite{hall2021professional} that task performance would not have significant differences across professions and set no condition on the occupation of participants. 
We had \hl{20} participants to complete the corresponding tasks for each of the five types of bar charts in the first phase, and \hl{20} participants for all three (type, parameter) pairs in the second phase.
Note that none of the participants acted repeatedly either in different phases or groups. Thus, we have \hl{$20 \times 5$ + $20$ = $120$} participants in total (ages between 18 and 45, 50 females).

\vspace{1.5mm}
\noindent\textbf{Procedure.}
We applied the following procedure in the lab study:
(i) explaining the tasks by the researcher, followed by training;
(ii) performing the study; and
(iii) a short interview.
Three training trials were provided to help participants get familiar with the task and our user study system. For each type of bar chart, we selected 5 random images without perturbation and then generated 5 images for each one of the 8 visual parameters by perturbing the corresponding parameter of those selected images, resulting in $5 + 5\times 8 = 45$  stimuli shown to each participant in the first phase.
In the second phase, \hl{each participant needed to complete 5 trials for each level of perturbations to three visual parameters identified as most influential in the previous phase, \ie $5 + 5\times (6+6+2) = 75$ stimuli.} In the interviews, we asked participants how they performed the task and which factors influenced their estimation through a questionnaire. Overall, it took about 15 minutes for each participant to finish the whole study, including 10 minutes for completing the task and 5 minutes for interview.

\begin{figure}[!t]
	\centering
	\includegraphics[width=\columnwidth]{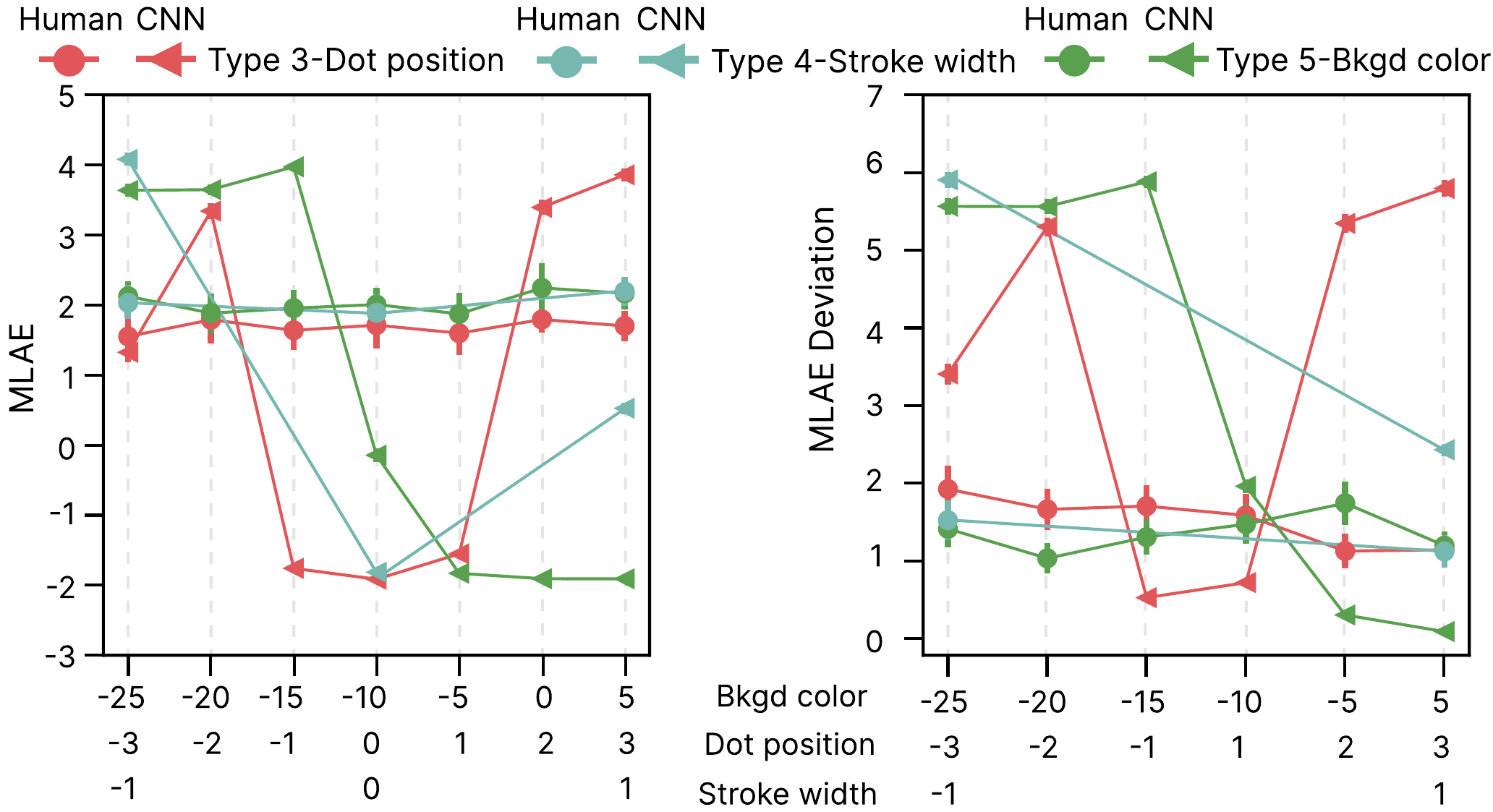}
	\caption{Comparison of bar chart tasks produced by humans and CNNs: The curves show how the mean MLAE values (a) and mean MLAE deviations (b) vary along the perturbation levels on the eight parameters.}
	\label{fig:human-cnn-comparison-prog}
 \vspace{-3mm}
\end{figure}

\vspace{1.5mm}
\noindent\textbf{Results Analysis.}
We first conducted a preliminary analysis of our collected data.
We used the D'Agostino-Pearson test to quantitatively confirm the normality of log errors' distribution. The results show that the log error distributions of human and CNN results are not normal distributions. Thus, following prior studies~\cite{cleveland1984graphical,haehn2019evaluating},
we compute the bootstrap distribution of the means of log errors using the 95\% confidence interval.
Note that the results from the between-subject study are treated as independent groups and do not support cross-group analysis.

\begin{figure*}[!t]
	\centering
	\includegraphics[width=0.9\textwidth]{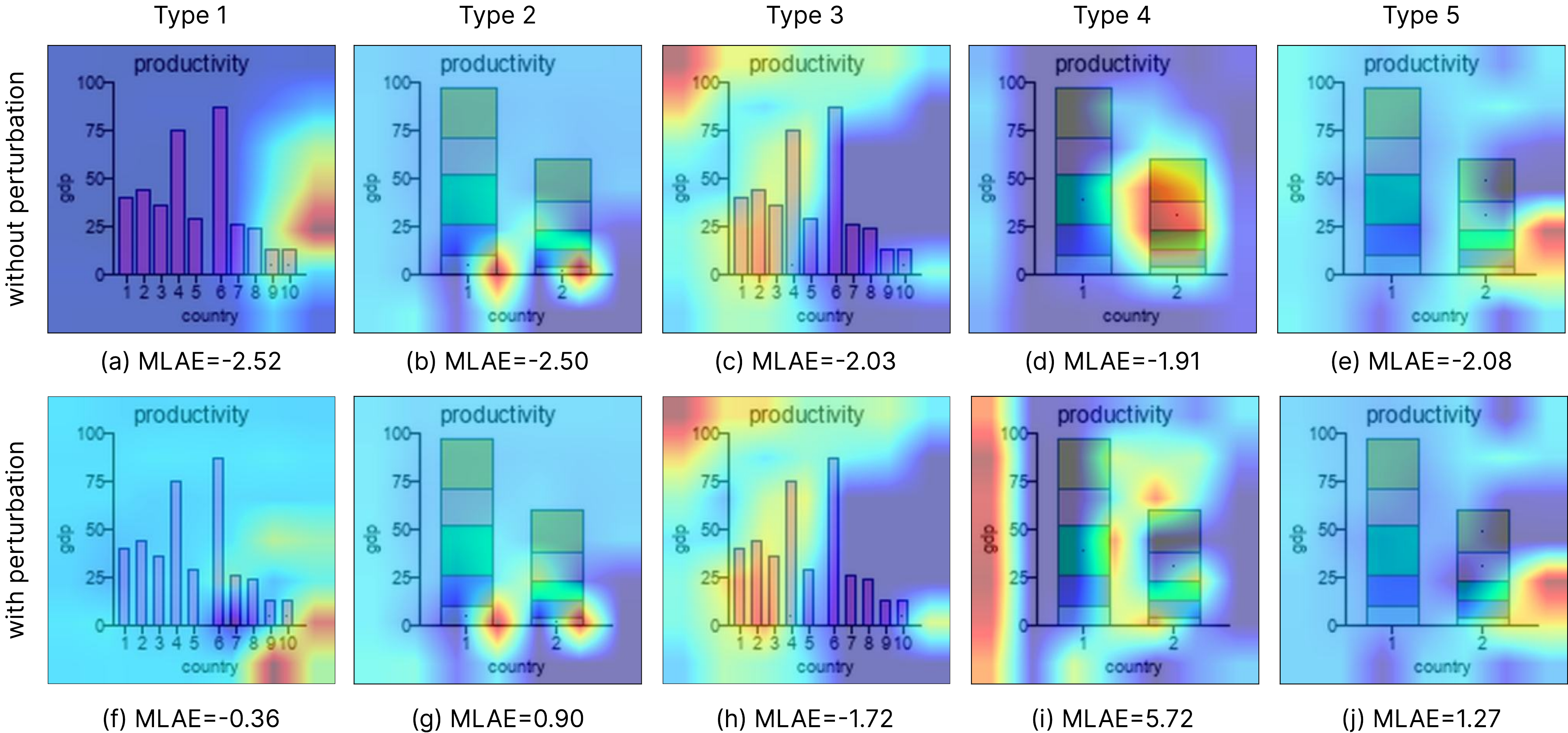}
	\caption{The Grad-CAM maps show the important regions in the chart image for estimating length ratios across five bar chart types. The two rows correspond to the charts without perturbations (first row) and with minor perturbations of the bar width (second row).}
	\label{fig:grad-cam}
    \vspace{-3mm}
\end{figure*}

Fig.~\ref{fig:human-cnn-comparison} summarizes the results of humans and CNNs for five bar chart tasks without and with the largest perturbations, respectively. We see that the MLAE values of humans are significantly larger than the ones of CNNs for all parameters without perturbation ($p<0.01$). Yet, the MLAE values of humans are smaller than the ones of CNNs for some parameters (\eg stroke color) with perturbations in the OOD setting. We make similar observations in Fig.~\ref{fig:human-cnn-comparison-prog}(a). Thus, we partially accept H1.
Fig.~\ref{fig:human-cnn-comparison-prog}(b) presents the summarized average MLAE deviation of CNNs and humans on all bar charts. We see that \textbf{humans are more robust than CNNs with smaller MLAE deviation for all tested levels of the most influential parameters}. Thus, we accept H2.

By using the Student's t-test to detect significant differences among parameters, we found that the human MLAE deviation shows significant differences for the bar length compared to the other parameters ($p<0.05$), whereas the MLAE deviation of the CNNs shows significant differences between all pairs of parameters ($p<0.01$). These statistics indicate that humans and CNNs might use different reasoning processes. Thus, we speculate that \textbf{humans are mainly influenced by the bar length but CNNs are influenced (and disturbed) by many other factors.}

This speculation is also supported by the interview results. Most participants mentioned that they did not notice the perturbations of parameters. They said it was hard to estimate the ratio when the difference between two bar lengths was too large or too small. These observations are consistent with the results produced by humans on the bar charts encoded within the value range $[10,93]$ and the ones in the range $[1,9] \cup [94,100]$, where the latter ones result in significantly larger MLAE values.

\section{Understanding and Improving CNNs} \label{sec:attention-cnn}

Our user study results show that humans tend to estimate bar length ratios based on specific parameters of the target bars while ignoring other task-irrelevant aspects of the chart. In contrast, CNNs' inference involves many visual parameters, which may explain why their performance is not as robust as humans.
To visually explore the inference of a CNN model, we use the gradient-weighted class activation mapping (Grad-CAM)~\cite{selvaraju2017grad} to localize the important regions of the chart image during inference.

\vspace{1.5mm}
\noindent\textbf{Grad-CAM Map.}
Since the last convolutional layers of the CNN model are expected to retain the task-relevant spatial information, we visualize their corresponding Grad-CAM maps. As shown in Fig.~\ref{fig:grad-cam}, the two rows of Grad-CAM maps correspond to the bar charts without and with the perturbation of bar width, respectively. We see that only type 2 focuses on the target bars (Fig.~\ref{fig:grad-cam}(b,g)) whose indices are fixed. We speculate this is caused by the sufficient capacity of the model in memorizing the indices information of the training data. In contrast, almost all regions of interest in the other types are unrelated to the task. For example, the focus region in type 1 is unrelated to all bars (Fig.~\ref{fig:grad-cam}(a,f)). 
Comparing the maps without and with perturbations, we see that even a small amount of perturbations on some visual parameters can dramatically change the resulting Grad-CAM maps.

\begin{figure}[t]
	\centering
	\includegraphics[width=\columnwidth]{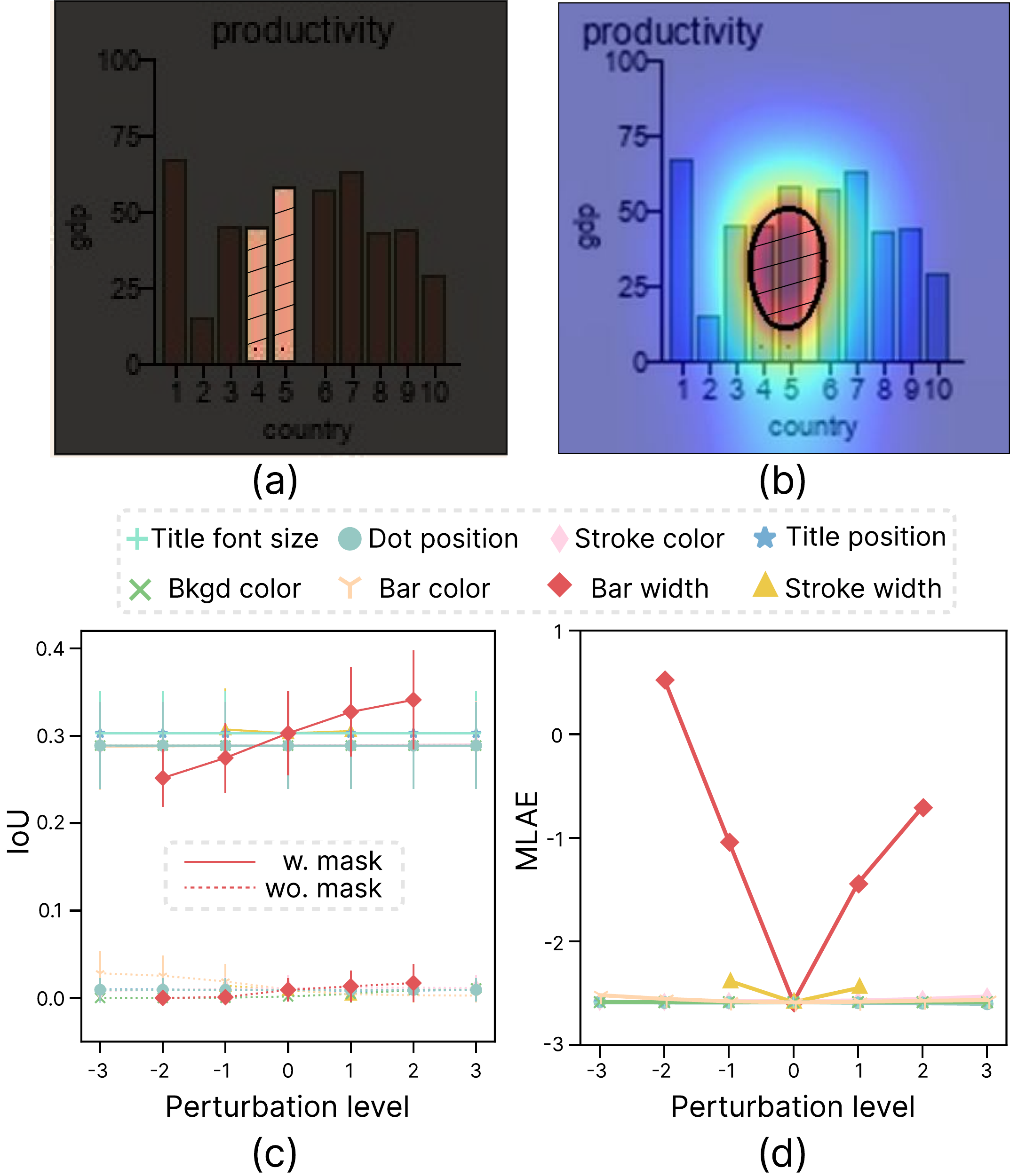}
	\caption{Improving the generalization performance of CNNs by providing segmentation masks. (a) The blended image of a chart and its segmentation mask, in which the area of target bars is shaded;
 (b) A perturbed chart blended with its Grad-CAM map, where the high-intensity region is shaded;
 And curves showing how IoU scores (c) and MLAE values (d) change over different levels of perturbations of one of eight parameters.
}
	\label{fig:cnn-improve}
 \vspace{-3mm}
\end{figure}

\vspace{1.5mm}
\noindent\textbf{Quantitative Analysis.}
We further assess the target region localization ability of CNNs by measuring the Intersection over Union (IoU) between the area of target bars $T$ and the high-intensity region of the Grad-CAM map $G$ as follows:
\begin{align}
    \text{IoU} = \frac{\left |T\cap G\right |}{\left |T\cup  G\right |}. \label{equ:iou}
\end{align}
We programmatically identify the segmented area of target bars (shown in Fig.~\ref{fig:cnn-improve}(a)), and extract the high-intensity regions of the Grad-CAM map (illustrated in Fig.~\ref{fig:cnn-improve}(b)) by empirically applying a threshold of $200$.
We compute the IoU scores for all type-1 stimuli in both IID and OOD tests. 
The results, visualized as dashed lines in Fig.~\ref{fig:cnn-improve}(c), show that mean IoU scores across all conditions are close to zero, indicating minimal overlap between the two areas.
This suggests that \textbf{CNNs might be using regions other than the target bars for relational reasoning on bar charts assuming Grad-CAM accurately reveals important features.}
However, there is no consensus on the effectiveness of different explainable AI methods~\cite{ibrahim2023explainable} (see results in the supplemental material), even though Grad-CAM remains a popular technique.

\begin{figure*}[t]
	\centering
	\includegraphics[width=0.85\textwidth]{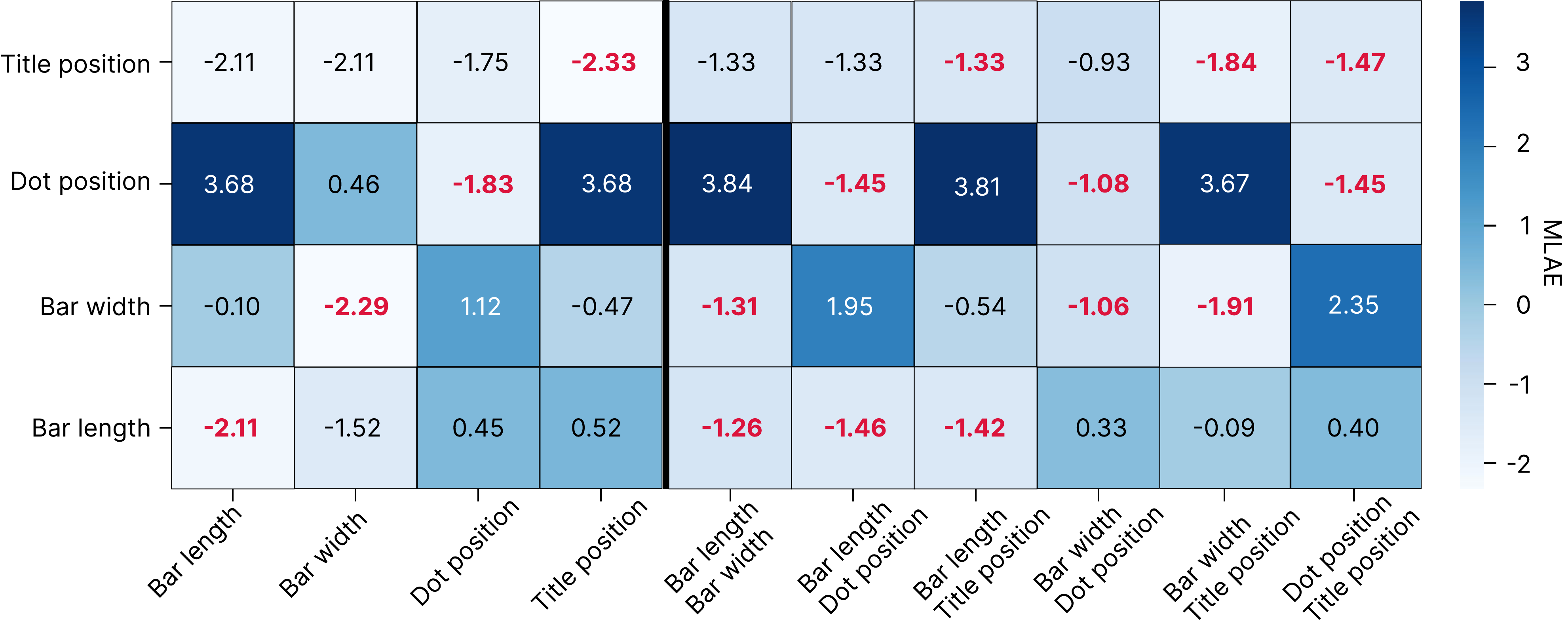}
	\caption{Improving the generalization performance of CNNs by augmenting training data with different perturbations. The heatmap shows the average MLAE values for testing CNNs on the chart images with different perturbations, where columns correspond to differently trained networks and rows show different test conditions. The cell highlighted by a bold red font indicates that the testing stimuli have similar perturbations as the ones in the training stimuli.}
	\label{fig:data-augmentation}
    \vspace{-3mm}
\end{figure*}

\vspace{1.5mm}
\noindent\textbf{Data Augmentation.}
To learn how well a CNN model trained with perturbed bar charts deals with other perturbations, we performed data augmentation using different combinations of the perturbations and then trained the model from scratch. When training a CNN model on two types of parameter perturbations (\eg bar length and bar width),
each type of perturbation was drawn uniformly. Taking the bar charts with the perturbation of bar length and bar width as an example, the bar length in each chart was encoded by the data range [1,100] and the bar width was a random value of the set \{4,5,6,7,8\}. Fig.~\ref{fig:data-augmentation} visualizes the testing result of the model on charts with the same or different perturbations.

We find that while all models are robust to the perturbation of title positions and those they were explicitly trained on, they all exhibit poor generalization performance on the perturbations that were not included in their training sets, especially the ones of the dot position. Notably, dot position and bar width seem to be the most influential parameters. Given the endless range of possible perturbations, it is impractical to train on every variation. Thus, we conclude that \textbf{data augmentation cannot close the human-CNN generalization performance gap in the task of relational reasoning with bar charts}.

\vspace{1.5mm}
\noindent\textbf{Segmentation Mask.}
In the second attempt to improve CNN performance, we add a segmentation mask of the target bars as an additional alpha channel to each chart image.
The segmentation mask is a binary mask of equal size to the original image, where pixels in the segmented foreground are all white and the background is all black.
We render these segmentation masks for all bar charts in the GRAPE dataset.
After training a CNN model with the corresponding RGB-$\alpha$ images, we tested the target region localization by Equation~\ref{equ:iou} and the generalization performance of the model as in Section~\ref{sec:generalization-performance}, the results are shown in Fig.~\ref{fig:cnn-improve}(c,d). We see that
(i) CNNs can better localize the target bars with the provided segmentation masks, as evidenced by the significant increase of the IoU scores, and (ii) CNNs' robustness has improved against most perturbations in visual encodings of bar charts such as title position, background color, and bar color.

However, this mask-enhanced CNN model is still sensitive to bar width and stroke width as shown in Fig.~\ref{fig:cnn-improve}(d). This suggests that \textbf{while segmentation masks may address the generalization issues related to texture/color in bar charts, generalizing to \emph{shape} variations remains a difficult task for CNNs.} This observation is consistent with the findings reported in~\cite{geirhos2018imagenet} that CNNs rely heavily on texture information when performing object recognition tasks and that models trained with stronger shape bias can exhibit greater robustness. Moreover, the resulting mean MLAE value is 2.28 for the bar lengths encoded in the range $[1,9] \cup [94,100]$, indicating that this model does not only use bar lengths to estimate ratios. In other words, even though the CNNs are provided with a segmentation mask, they cannot estimate length ratios as well as humans in certain cases.

Because the segmentation masks are based on the complete and accurate segmentation of the target bars, removing the pixel dot in the RGB image (see the dots in Fig.~\ref{fig:cnn-improve}(a)) does not degrade the model's performance. Yet, the CNNs cannot complete the task when the mask is incomplete. For example, if we only denote the pixel dot in the alpha map, the training process does not converge at all. Even replacing the pixel dot with a small Gaussian blob results in an MLAE value of 4.32.

\section{Discussion}

\vspace{1.5mm}
\noindent\textbf{Evaluating Vision Models with Grammar of Graphics.}
For over a decade, the computer vision community has been using ImageNet~\cite{deng2009imagenet}, a large-scale dataset that contains 3.2 million natural images, as the benchmark dataset for training and evaluating the generalization performance of vision models.
Compared with natural images, the design space of data visualization contains more complex and hierarchical semantic information. Each visual element in visualizations has its own semantic meaning that contributes to a specific visual expression. For example, even the smallest components of a diagram, such as tick marks, are important indications of the correspondence between numbers and the coordinate axis. This makes it harder for neural networks to learn appropriate features and relations between different visual marks.
Synthesizing large-scale visualization datasets using the grammar of graphics, such as GRAPE,
will allow the research community to conduct more systematic evaluations of graphical perception of machines and humans.

\vspace{1.5mm}
\noindent\textbf{Network Architecture and Task-oriented Attention.}
Our experiments show that title position affects the robustness of CNNs on certain bar charts, even though it is irrelevant to graphical perception tasks.
In contrast, humans do not have this issue due to our robust attention mechanism. 
The Transformers architecture~\cite{vaswani2017attention} uses self-attention mechanism that has proven highly effective in~\cite{dosovitskiy2020image,liu2021swin} and may be robust to the perturbations on certain parameters.
Since visualizations are often task-dependent, different tasks have different relevant visual encodings. For example, reading values from a bar chart requires understanding the axis, while estimating the ratio between bars requires comparing bar lengths. Hence, we believe that a task-oriented attention mechanism that can understand and reason about task-related visual encodings will enhance performance in graphical perception and computational visualization analysis.

\section{Conclusions and Future Work}
In this paper, we presented a comprehensive generalization analysis for CNNs \hl{and humans} on graphical perception tasks with bar charts. We find that appropriately trained CNNs can outperform humans on ratio estimation tasks. Based on the carefully selected configuration (ResNet152 and AdamW solver), we show that CNNs also outperform humans on graphical perception tasks using standard visualizations specified by Vega-Lite where the visual parameter space is greatly expanded compared to previous studies~\cite{cleveland1984graphical,haehn2019evaluating}.

Furthermore, we compared the generalization performance of CNNs and humans by progressively perturbing the design space of the standard visualizations and conducted a user study. We created a large dataset of visualizations called GRAPE by programmatically exploring the grammar of graphics using Vega-Lite. The results of our user studies show that (i) CNNs outperform humans but are less robust when the training data and test data are similar in visual encodings (IID), and (ii) human performance is more robust to perturbations of various visual parameters than that of CNNs (OOD). Hence, we conclude that CNNs' inference process involves many visual variables, yet they lack robust recognition of the target visual marks. Note that the human performance is collected from the user study with a small number of participants,  we will conduct a larger-scale study to further confirm this conclusion in the future.

This work is the first step in the systematic study of the generalization behavior of CNNs for visualization applications. There is still a lot to explore. First, we identified our hyper-parameters in an ad-hoc way. It would be better to use recently proposed AutoML techniques~\cite{he2021automl} to automatically build and train optimal neural networks for graphical perception tasks.
Second, our analysis results reveal that some factors like background color and stroke color that are unrelated to the graphical perception tasks heavily impact the OOD generalization of CNNs.
We plan to explore more recently proposed deep learning frameworks, such as Vision Transformers~\cite{dosovitskiy2020image} and CLIP~\cite{radford2021learning}, to see if they are able to close the gap between humans and CNNs for visualization graphical perception tasks.
Lastly, there are many visualizations (\eg infographics~\cite{otten2015infographics,chen2019towards}) that do not comply with the grammar of graphics and still effectively communicate information. We would like to examine the CNNs' graphical perception abilities on such creative visualizations. 

While this work focuses on graphical perception tasks, we believe that many applications regarding data visualizations should be evaluated from the generalizability perspective. 
For example, building on the recent success of LLMs, many chart-centered large multimodal models (LMMs)~\cite{meng2024chartassisstant,liu2023mmc} have been developed.
While these models demonstrated strong performance on several chart QA benchmarks, their sensitivity to various visual variations may still fall short of matching human-level robustness in relational reasoning tasks.
Our study highlights the necessity of evaluating the OOD generalization of such LMMs.
In the future, we plan to create a research benchmark for evaluating the generalization performance of different visualization techniques and analysis tasks.

\section*{Acknowledgments}
This work was partially supported by the grants of the NSFC (No.62132017), the Shandong Provincial Natural Science Foundation (No.ZQ2022JQ32), the Fundamental Research Funds for the Central Universities, the Research Funds of Renmin University of China, and the Singapore Ministry of Education (MOE) Academic Research Fund (AcRF) Tier 2 (No.T2EP20222-0049).

\bibliographystyle{abbrv-doi-hyperref}
\bibliography{references}

\vfill

\end{document}